\documentclass[a4paper, 10 pt, conference]{ieeeconf}

\IEEEoverridecommandlockouts                              

\usepackage{times}
\usepackage{soul}
\usepackage{url}
\usepackage[utf8]{inputenc}
\usepackage{amsmath,amssymb}
\DeclareMathOperator{\E}{\mathbb{E}}
\usepackage{graphicx}
\usepackage{siunitx} 
\usepackage{subcaption} 
\usepackage{color}
\usepackage[noadjust]{cite}

\newcommand{\etal}{\textit{et al.}}
\newcommand{\eg}{\textit{e.g.,}}
\newcommand{\ie}{\textit{i.e.,}}

\title{\LARGE \bf
MCENET: Multi-Context Encoder Network for Homogeneous Agent Trajectory Prediction in Mixed Traffic}

\author{Hao Cheng$^{1,*}$, Wentong Liao$^{2,*,\dag}$, Michael Ying Yang$^{3}$, Monika Sester$^{1}$ and Bodo Rosenhahn$^{2}$
\thanks{$^{1}$Hao Cheng and Monika Sester are with the Institute of Cartography and Geoinformatics, Leibniz University Hannover,
        Appelstr. 9a, 30167, Germany
        {\tt\small \{cheng, sester\}@ikg.uni-hannover.de}}%
\thanks{$^{2}$Wentong Liao and Bodo Rosenhahn are with the Institute of Information Processing, Leibniz University Hannover,
        Appelstr. 9a, 30167, Germany
        {\tt\small \{liao, rosenhahn\}@tnt.uni-hannover.de}}%
\thanks{$^{3}$Michael Ying Yang is with the Scene Understanding Group, University of Twente, Netherlands
        {\tt\small michael.yang@utwente.nl}}%
\thanks{*~These authors contributed equally to this work.}%
\thanks{\dag~Corresponding author.}%
}

\begin{document}

\maketitle
\thispagestyle{empty}
\pagestyle{empty}

\begin{abstract}
  Trajectory prediction in urban mixed-traffic zones (a.k.a. shared spaces) is critical for many intelligent transportation systems, such as intent detection for autonomous driving.
  However, there are many challenges to predict the trajectories of heterogeneous road agents (pedestrians, cyclists and vehicles) at a microscopical level. For example, an agent might be able to choose multiple plausible paths in complex interactions with other agents in varying environments. 
  To this end, we propose an approach named \emph{Multi-Context Encoder Network} (MCENET) that is trained by encoding both past and future scene context, interaction context and motion information to capture the patterns and variations of the future trajectories using a set of stochastic latent variables.
  In inference time, we combine the past context and motion information of the target agent with samplings of the latent variables to predict multiple realistic trajectories in the future.
  Through experiments on several datasets of varying scenes, our method outperforms some of the recent state-of-the-art methods for mixed traffic trajectory prediction by a large margin and more robust in a very challenging environment. The impact of each context is justified via ablation studies.
\end{abstract}

\section{Introduction}
\label{sec:introduction}
\begin{figure}[t]
\begin{center}
 \includegraphics[clip=true,trim=0pt 0pt 0pt 2in, width=1.0\linewidth]{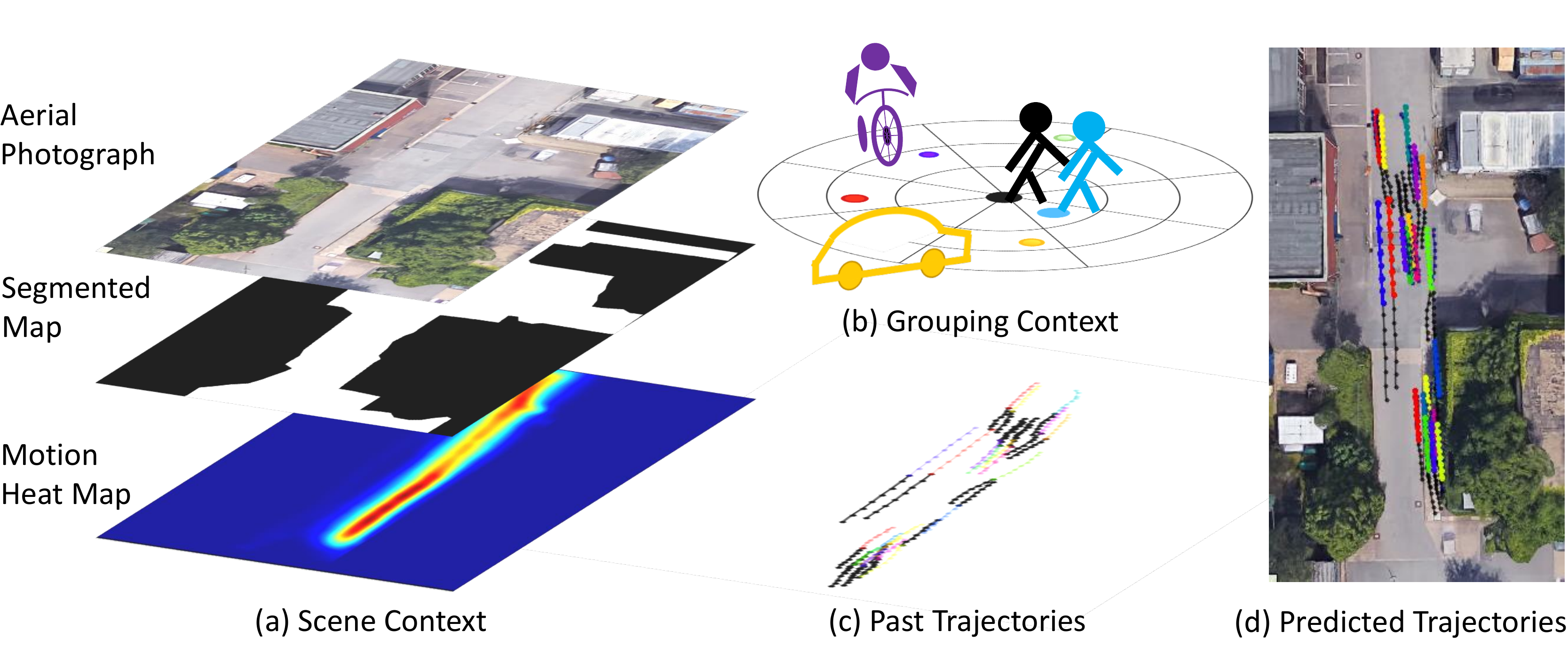}
\end{center}
   \caption{\small{Predicting the future trajectory (d) by observing the past trajectories (c) considering the scene (a) and grouping context (b). Three kinds of scene context: (1) aerial photograph provides overview of the environment, (2) segmented map defines the accessible areas respective to road agents' transport mode and (3) the motion heat map describes the prior of how different agents move. Different colors denote different agents or agent groups.}}
\label{fig:example}
\vspace{-12pt}
\end{figure}

Correctly understanding the motion behavior of road agents in the near future is crucial for many intelligent transportation systems (ITS), such as intent detection \cite{goldhammer2013early,hashimoto2015probabilistic,koehler2013stationary}, trajectory prediction \cite{alahi2016social,gupta2018social,vemula2018social} and autonomous driving \cite{franke1998autonomous}, especially in urban mixed-traffic zones (a.k.a. shared spaces \cite{reid2009dft}). 
Trajectory prediction is defined as to predict the plausible and social-acceptable future trajectories of target agents by observing their history trajectories, as shown in Fig.~\ref{fig:example}(c)(d).

At a microscopical-level coordinate system at, \eg~each half second in the future, the above task can be extremely difficult due to the mutual effects from three major factors: \textit{ego motion}, \textit{interaction} and \textit{environment}. To be more specific, 
(1) The same kind of agent is likely to behave differently in varying environments because of different scene context and each agent has more than one plausible future paths.
(2) The dynamics in interactions among agents as well as between single agent and group agents are complex.
(3) Interaction in shared space is full of uncertainties. Contrary to conventional traffic design where road resources are allocated to road users by time or space segregation, shared space largely removes road signs, signals, and markings, forcing direct interaction between mixed traffic participants.

There are many works trying to cope with the above challenges in different aspects: single agent-to-agent interaction \cite{robicquet2016learning,alahi2016social,gupta2018social}, single and group interaction \cite{yamaguchi2011you,rinke2017multi,bisagno2018group,cheng2019pedestrian}, agent-to-environment interaction \cite{burstedde2001simulation,kitani2012activity}, considering both interaction and environmental factors but only for homogeneous agents (\eg~pedestrians) \cite{yamaguchi2011you,xue2018ss,xu2018encoding}. 
Recently, more and more works address the problems of trajectory prediction by generating multiple paths \cite{lee2017desire,gupta2018social,sadeghian2018sophie,zhao2019multi} and generalize the task for mixed traffic \cite{schonauer2012modeling,rinke2017multi,cheng2018modeling,cheng2018mixed,chandra2019traphic}. 
However, it lacks work that comprehensively tackles the aforementioned challenges within one framework for mixed traffic multi-path trajectory prediction.

\begin{figure}[t!]
	\centering
	\includegraphics[clip=true,trim=0pt 0pt 0pt 0pt, width=1\linewidth]{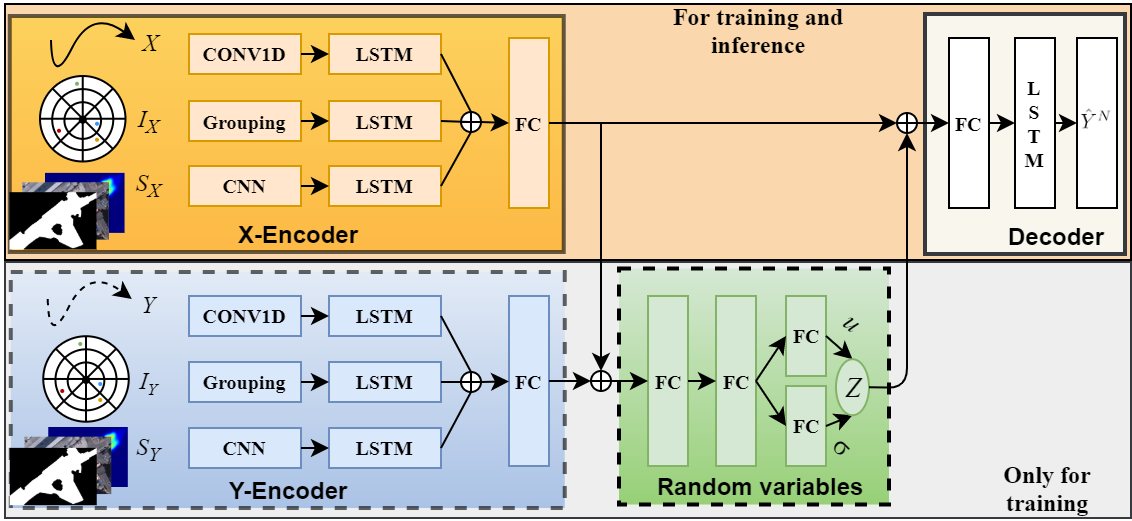}
	\caption{\small{The pipeline for the proposed method. The ground truth $Y$ and the associated interaction $I_Y$ and scene context $S_Y$ are injected to the input only in training. They are not available in inference. The latent variables $z$ are sampled $N$ times and concatenated with the output of X-Encoder for predicting multiple future paths.}}
	\label{fig:pipeline}
\vspace{-12pt}
\end{figure}

To fill up the research gap, we propose \emph{Multi-Context Encoder Network} (MCENET) that predicts multi-path trajectories of heterogeneous road agents by introducing grouping and scene contexts. An overview of our framework is depicted in Fig.~\ref{fig:pipeline}.
MCENET consists of two encoders and a decoder: an encoder is trained to encode the past information including the motion and context information of target agent while the other encoder is for the future information.
Then, the two encoded information are fused and then forwarded to learn a latent space that describes the distribution of the future trajectories.
The decoder is trained to predict multi-path trajectories of target agent depending on its past information and a set of stochastic latent variables which are sampled from the learned latent space.
For each module an LSTM is trained to encode/decode the sequential information separately. In Sec.~\ref{sec:method} we will discuss our method in detail.

The innovations of our method are summarized as follows:
\begin{itemize}
\item[1] \textbf{Grouping Context.} Each agent's is affected by other agents around it, \eg~a person will have the similar motion of others within a group. Therefore, distinguishing the group and non-group agents for a target agent is useful for analyzing its motion.
\item[2] \textbf{Scene Context}. Agents' behaviors are constrained by the environment, such as space layout and building deployment, especially in a shared space. To explore the effect from the scene, 
three kinds of scene context are studied in this work: the \emph{motion heat maps} describe the prior of how different agents move; \emph{aerial photography images} provide global visual information over the scene; and the \emph{segmented maps} define the accessible areas respective to road agents' transport mode.
\item[3] \textbf{Multi-path Trajectories.} Given a past trajectory, there are more than one plausible future paths. Our work focuses on predicting multiple plausible and socially-acceptable trajectories.
\item[4] \textbf{Heterogeneous road agents.} We analyze pedestrians, cyclists and vehicles rather than only consider a specific kind of agents (pedestrians or cars) which is normally done by previous works \cite{yamaguchi2011you,yi2016pedestrian,lee2017desire,xue2018ss,xu2018encoding,sadeghian2018sophie}. 
\end{itemize}
Our approach is validated on several datasets and outperforms the recent methods. The impact of each proposed module in our framework is justified via ablation studies. 
The code of our method is available at~\url{https://github.com/haohao11/MCENET}

\section{Related Work}
\label{relatedwork}
Trajectory prediction has been attracting attention in ITS for decades. In general, the approaches can be categorized into two branches: expert models with hand crafted rules \cite{helbing1995social,burstedde2001simulation,yamaguchi2011you,kitani2012activity,schonauer2012modeling} and data--driving, especially deep--learning (DL) models with different representation methods \cite{lee2017desire,gupta2018social,xue2018ss,xu2018encoding,sadeghian2018car,sadeghian2018sophie,zhao2019multi,chandra2019traphic,liang2019peeking}.

\textbf{Expert Models.} The Social Force Model (SFM) is one of the most well-known approaches for pedestrian agent simulation in crowded space, which uses different forces on the basis of classic physic mechanics to mimic human behavior \cite{helbing1995social}. The repulsive force prevents the target agent from colliding with others or obstacles and the attractive force drives the agent close to its destination or companies. Extra forces are extended to model more complex interactions \cite{robicquet2016learning} and mixed traffic \cite{rinke2017multi}.
Cellular Automata models divide environment space into small identical discrete cells. The movement of agents is governed by a set of manually defined rules in those cells \cite{burstedde2001simulation}.
Hidden variable Markov decision processes are used to model agent-to-environment interaction \cite{kitani2012activity}.
The Energy function is proposed to model pedestrian behavior with the consideration of personal, social and environmental factors \cite{yamaguchi2011you}. 
Game Theory is used for simulating the complex interactions in mixed traffic of heterogeneous agents \cite{schonauer2012modeling}.

However, designing good rules for the expert models is complex and requires professional knowledge. Meanwhile, those models have difficulties in scaled-up problems (\eg~large number of agents) when the rules are no longer applicable (\eg~structural alteration of the space).

\textbf{Deep Learning Models.} To overcome these drawbacks of expert models, many recent works \cite{alahi2016social,gupta2018social,lee2017desire,xue2018ss,xu2018encoding,sadeghian2018car} resort to the deep learning technologies \cite{lecun2015deep}, which are able to learn powerful representation from large-scale data. 

DL models are used to automatically learn interactions between agents. For instance, Social-LSTM proposed in \cite{alahi2016social} uses a social pooling layer to capture the interactions between a target agent and individual neighborhood agents in a pre-defined interactive zone. 
Nevertheless, it does not consider the grouping context. When a neighborhood agent is a company of the target agent, it is treated the same as the other neighborhood agents that have no social connection with the target one. In reality, pedestrians in a group may behave differently than individual pedestrians. In the group, pedestrians tend to synchronize their speed and maintain a certain distance for communication and visibility between each other \cite{yamaguchi2011you,rinke2017multi}. To this end, grouping is incorporated in \cite{bisagno2018group,cheng2019pedestrian} to differentiate the repulsive and attractive effects explicitly for group and non-group members.

Many works consider the interactions between agents but ignore the impact of the environment. The scene context of the space (\eg~buildings or trees) may constrain certain movements. A recent model called SS-LSTM \cite{xue2018ss} reports better performance for pedestrian trajectory prediction by exploring scene information from aerial photographs, which provides global context for understanding the environment. Scene context has been proven to be beneficial for trajectory prediction in many recent studies \cite{manh2018scene,bartoli2018context,sadeghian2018car,liang2019peeking,zhao2019multi}. 

However, most of the aforementioned methods only predict one future trajectory based on each history information for an agent. There might be multiple plausible paths that an agent could take in the future. 
For example, an agent can have some degrees of freedom to move in crowd with slightly different speed and orientation.
To generate multiple plausible trajectories of the target agent, generative models are introduced into this task. Social generative adversarial network (S-GAN) proposed by \cite{gupta2018social} trains a generator to generate future trajectories from noise. Meanwhile a discriminator is trained to judge whether the generated one is fake or not. The performance of the two modules are enhanced mutually and the generator is able to generate trajectories that are precise as the real ones. Conditional variational autoencoder (CVAE)~\cite{kingma2013auto,kingma2014semi} is proposed to predict multiple plausible trajectories in \cite{lee2017desire}. CVAE is trained to learn the latent stochastic space of the future trajectory depending on the past information. Multiple trajectories of the target agent are predicted from its history motion by introducing a set of stochastic latent variables.

Most of the previous works focus on predicting trajectory of a specific kind of agents (mainly pedestrians). However, the real-world urban traffic scenarios are more complex and there are different kinds of participants (pedestrians, cyclists and vehicles). A hybrid architecture is proposed in \cite{chandra2019traphic} that combines convolutional neural networks (CNN) with LSTMs to encode trajectory information and different dynamic constraints \eg~agent shape, velocity and traffic concentration,  for trajectory prediction of heterogeneous road agents. 
Cheng \etal~\cite{cheng2018mixed} propose to incorporate field-of-view to distinguish different transportation modes and Cheng \etal~\cite{cheng2018modeling} map collision probability for different types of road agents in mixed traffic trajectory prediction.

\section{Methodology}
\label{sec:method}
\subsection{Problem Definition}
\label{subsec:definition}
The multi-path trajectory prediction problem is defined as: for an agent $i$, received as input its observed trajectories $\mathbf{X}_i=\{X_i^1,\cdots,X_i^T\}$ and predict its $n-th$ plausible future trajectory $\hat{\mathbf{Y}}_{i,n}=\{\hat{Y}_{i,n}^1,\cdots,\hat{Y}_{i,n}^{T'}\}$. $T$ and $T'$ denote the sequence length of the past and being predicted trajectory positions, respectively. The trajectory position of $i$ at time step $t$ is characterized by the coordinate as $X_i^t=({x_i}^t, {y_i}^t)$ (3D coordinates are also possible, but in this work only 2D coordinates are considered) and so as $\hat{Y}_{i,n}^{t'}$.
For simplicity, we omit the notation of time steps in the following parts of the paper.
The objective is to predict its multiple plausible future trajectories $\hat{\mathbf{Y}}_i = \hat{\mathbf{Y}}_{i,1},\cdots,\hat{\mathbf{Y}}_{i,N}$ that are as accurate as possible to the ground truth $\mathbf{Y}_i$. This task is formally defined as $\hat{\mathbf{Y}}_i^n = f(\mathbf{X}_i, A, S), ~n \in N$. The total number of the predicted trajectories is denoted by $N$. 

\subsection{Input Information}
\label{subsec:input}
\paragraph{Motion Information} Specifically, we use the offset $({\Delta{x}_i}^t, {\Delta{y}_i}^t)$ of the trajectory positions between two consecutive time steps as the motion information instead of the coordinates within the image. Because the offset is independent from the given image and can be interpreted as speed over time steps that are divided by a constant duration. 
As long as the original position is known, the sequence of offsets can be converted back to positions by a cumulative summation function.
Because different kinds of agent have different motion patterns, we adopt one-hot representation to indicate the type of agent explicitly and concatenate it with the motion representation.

\paragraph{Grouping Context} We use a polar occupancy grid to parse the interaction context between the target agent and its neighborhood agents, which is widely adopted~\cite{lee2017desire,xu2018encoding}. The polar grid is divided into a certain number of cells which are sorted according to the orientation and distance to the centroid (see Fig.~\ref{fig:example}(b)). Hence, each cell represents a unique position to the centroid. Analogously, a neighborhood is mapped to a cell using the orientation and distance to the target agent (centroid) at each time step. 

With the consideration of grouping context for distinguishing the effect of group and non-group members on the target agent, on top of the occupancy, Density-Based Spatial Clustering of Applications with Noise (DBSCAN) \cite{ester1996density} is utilized with time span for group detection. At each step during the observation time, present agents are clustered. The minimum number of points (MinPts) is set to $2$ because a group (cluster) contains at least two agents. A pre-defined threshold is set to the maximum Euclidean distance $\epsilon$ from neighborhood point to the core points in the DBSCAN cluster. A neighborhood agent is defined as a group member for the target agent if they co-exist in the same cluster over a certain rate of the observed time steps. When a neighborhood agent is detected as a ``friend'' of the target agent, it will not be stored in the grid cell and is not treated as obstacles with repulsive effect on the target agent.

The process of occupancy and grouping is formulated as:
\begin{equation}
\label{eq:occupancyandgrouping}
    \text{cell}_m^{t}(r, d) = {\sum} 1 [{j \in B_i^{t} \,\text{and} \,j \not\in G_i}], \, j\neq i
\end{equation}
$\text{cell}_m^{t}(r, d)$ stands for the $m$-th cell grid with orientation $r$ and distance $d$ to the centroid at time $t$. 
For agent $i$, $B_i^{t}$ and $G_i$ stand for the set of neighbors at time $t$ and the set of group members, respectively. If there is a non-group member in $\text{cell}_m^{t}(r, d)$, its value will be added by one. 

\paragraph{Scene Context}
Three types of scene context are studied for trajectory prediction: heat maps, aerial photographs and segmented maps (see Fig.~\ref{fig:example}(a)). 

\textbf{Heat maps} are statistic distribution of the trajectories in the training dataset. With the assumption that road users tend to follow the trajectories of others, the areas visited more often in the past are more likely to be visited in the future. 
Hence, we generate heat maps for each type of agents and use a Gaussian filter with a large kernel to expand the possible areas to the contiguity that can be covered in order to reduce the strong statistic bias.  

\textbf{Aerial photographs} are taken from the bird-eye's-view over the space to provide global context, such as deployments of buildings, trees, and streets.

\textbf{Segmented maps} are binary masks indicating the areas that can be accessed by road agents corresponding to their transport mode. White areas are accessible (\eg~road and sidewalk for pedestrians) and black areas are not accessible (\eg~buildings and trees for vehicles).  

\subsection{Multi-Context Encoder Network}
\label{subsec:contextencoderanddecoding}
MCENET is inspired by the structure of Conditional Variational Autoencoder (CVAE)~\cite{kingma2013auto,kingma2014semi}. CVAE is a generative model that uses a set of latent variables to encode the observed variables. During training, the label of the input variable is inserted into the input variable as the condition for learning a hidden space with a set of latent variables, which can follow, \eg~a Gaussian distribution. In inference, the latent variables can be sampled multiple times to reconstruct the input variable with some controlled variations. This mechanism of CVAE allows us to generate more than one outputs with only one input. There is no need to explicitly specify the structure of the output~\cite{doersch2016tutorial}.

In order to turn the problem of trajectory prediction into a generative reconstruction problem, we first encode future and past trajectories into a set of latent variables in training. Then the prediction can be treated as reconstructing the future trajectories depending on the past trajectory and the latent variables~\cite{lee2017desire}.   
\begin{equation}
\label{eq:CVAE}
\begin{split}
    \log{P(Y|X)} \geq - \text{KL}(Q(z|Y, X)||P(z)) \\ 
   + \E_{Q(z|Y, X)}[\log{P(Y|z, X)}].
\end{split}
\end{equation}
Eq.~\eqref{eq:CVAE} denotes the reconstruction process. $Y$ and $X$ stand for future and past trajectories, respectively, and $z$ for latent variables. The objective of Eq.~\eqref{eq:CVAE} is to maximize the conditional probability $\log{P(Y|X)}$, which is equivalent to minimize $\ell (\hat{Y}, Y)$ and at the same time minimize the Kullback-Leibler divergence. 
In order to enable back propagation for stochastic gradient descent in $\E_{Q(z|Y, X)}[\log{P(Y|z, X)}]$, a re-parameterization trick \cite{rezende2014stochastic} is applied to $z$, where $z$ can be re-parameterized by $z = \mu + \sigma \odot \epsilon$. $\epsilon$ also follows a Gaussian distribution. Both right terms in Eq.~\eqref{eq:CVAE} can be parameterized by neural networks. 

Whereas, as the name of \emph{MCENET} implies, we extend the CVAE structure to \textbf{M}ulti-\textbf{C}ontext \textbf{E}ncoder \textbf{NET}work for different types of information. Fig.~\ref{fig:pipeline} depicts the pipeline of the framework. X-encoder and Y-encoder encode past and future scene context, interaction context and motion information in parallel, in order to consider environment, interaction and motion factors alone the complete time horizon. 

X-encoder is used to encode the past information. First, the CONV-1D is used to learn motion features along the time axis, DBSCAN is used to detect agent groups in interactions, and the three-layer CNN is used to extract features of scene context. Alternatively, rather than training the CNN from scratch, the feature extractor can also be substituted by a pre-trained network, \eg~MobileNet~\cite{howard2017mobilenets}.
Then, the features of these three branches are fed to different LSTMs for learning the hidden information at each time step. The outputs of the LSTMs are concatenated and passed through a fully connected (FC) layer followed by the ReLU activation for fusing features. 
The output of X-encoder is denoted as $\Phi_X(.)$.
Y-encoder is used to encode the future information and works in the same way as X-encoder in parallel. The output of Y-encoder is denoted as $\Phi_Y(.)$.
During training, the encoded past and future representations $\Phi_X(.)$ and $\Phi_Y(.)$ are concatenated and forwarded to two FC layers followed by the ReLU activation. The following two FC layers are trained to learn the mean and variance of the distribution of the latent variables $z$, respectively.
In the end, $\Phi_X(.)$ and the sampled latent variables $z$ are concatenated and fed to the decoder to reconstruct $Y$. The decoder consists of a FC layer for fusion and dimension reduction and one LSTM for sequentially prediction.

During inference, Y-encoder is removed and the past information is encoded and fused by X-encoder in the same way as in the training stage. To generate a future prediction sample, the latent variable $z$ is sampled from $\mathcal{N}(\mu, ~\sigma^{2}*I)$ and concatenated with $\Phi_X(.)$ as the input of the decoder:
\begin{align}
\label{eq:encoding}
    z &= Q({{\Phi}_Y(.), \Phi}_X(.)), ~z \sim \mathcal{N}(\mu, ~\sigma^{2}*I),\\
\label{eq:decoding}
    \mathbf{\hat{Y}} &= P(\Phi_X(.), z).
\end{align}
This step is repeated N times to generate N samples of future prediction.
The MSE loss ${\ell}_2 (\mathbf{\hat{Y}}, \mathbf{Y})$ (reconstruction loss) and the $\text{KL}(Q(z|\mathbf{Y}, \mathbf{X})||P(z))$ loss are used to train our model.
The MSE loss will force the reconstructed results as close as possible to the ground truth and the KL-divergence loss will force the latent variables $z$ to be a Gaussian distribution.    

\subsection{Trajectories Ranking}
\label{subsec:ranking}

A bivariate Gaussian distribution is used to rank the multiple predicted trajectories $\hat{Y}^1,\cdots,\hat{Y}^N$ for each agent. At each time step, the predicted positions $({\hat{x}_{i,n}}^{t'}, {\hat{y}_{i,n}}^{t'})$, where $~n{\in}N$ at time step $t'\in T'$ for agent $i$, are used to fit a bivariate Gaussian distribution $\mathcal{N}({\mu}_{xy},\,\sigma^{2}_{xy}, \,\rho)^{t'}$. The predicted trajectories are sorted by the joint probability density functions $p(.~)$ over the time axis using Eq.~\eqref{eq:pdf}\eqref{eq:sort}. $\widehat{Y}^\ast$ denotes the most-likely prediction out of $N$ predictions.

\begin{align}
\label{eq:pdf}
    P({\hat{x}_{i,n}}^{t'}, {\hat{y}_{i,n}}^{t'}) \approx p[({\hat{x}_{i,n}}^{t'}, {\hat{y}_{i,n}}^{t'})|\mathcal{N}({\mu}_{xy},\sigma^{2}_{xy},\rho)^{t'}]\\
\label{eq:sort}
    \widehat{Y}^\ast = \text{arg\,max}\sum_{n=1}^{N}\sum_{t'=1}^{T'}{\log}P({\hat{x}_{i,n}}^{t'}, {\hat{y}_{i,n}}^{t'})
\end{align}

\section{Experiment}
\label{sec:experiment}

\subsection{Datasets}
\label{subsec:datasets}
We first validate our method for mixed traffic on the benchmark dataset Gates3 \cite{robicquet2016learning} and conduct extended experiments on the other two datasets HBS \cite{rinke2017multi} and HC \cite{cheng2019pedestrian}, which have different scenes to evaluate the generalization ability of our model. 
Gates3 is one of the most challenging subsets of the Stanford Drone Dataset \cite{robicquet2016learning}. It was captured from a very busy roundabout in Stanford. After removing some wrong trajectories, it contains $9.9k$ frames and 159 pedestrians and 223 cyclists.
HBS dataset was collected near a busy train station with pedestrian cross-walking among vehicles and cyclists. There are $3.6k$ frames and 115 pedestrians, 22 cyclists and 338 vehicles.
HC dataset was taken over a street with buildings and trees on both sides of a university campus. It has $3.5k$ frames with 384 pedestrians, 42 cyclists and 13 vehicles.
The frame rate of Gates3 has been down-sampled to \SI{2}{fps}, in order to keep it as consistent as the other two datasets.  
Each dataset has been split into training (last \SI{70}{\percent} of the total time steps) and test (first \SI{30}{\percent}) subsets. 
Conventionally, 8 steps of history trajectories are taken as observation and the next 8 steps are predicted.
Longer term prediction is possible, but \SI{2.4}{s} are sufficient for most human to respond to emergence \cite{taoka1989brake}. Hence, here we report performances for the next \SI{4}{s} prediction.

Besides the validation in mixed traffic, we also validate our method on pedestrian benchmark datasets ETH \cite{pellegrini2009you} and UCY \cite{lerner2007crowds}. These datasets contain various challenging interactive scenarios between pedestrians in different public spaces, such as single pedestrian vs. single pedestrian, single pedestrian vs. pedestrian group, pedestrian group vs. pedestrian group. In total, five sub-datasets (Eth and Hotel from ETH and Zara1, Zara2 and Univ from UCY) are selected. By default, the time step has been down-sampled to \SI{2.5}{fps}.
In order to make full use of the datasets for training models, we follow the prior works \cite{xu2018encoding,gupta2018social,sadeghian2018sophie} that use the leave-one-out cross-validation fashion---one dataset is for test and the rests are for training~\cite{alahi2016social}---and prediction time horizon---observing 8 steps and predicting 8 and 12 steps, respectively. 

\subsection{Evaluation Metrics}
\label{sec:evaluationmetrics}
The average displacement error (ADE) and the final displacement error (FDE) are the two most commonly applied metrics to measure the performance in terms of trajectory prediction~\cite{alahi2016social,gupta2018social,sadeghian2018sophie}. 
ADE is the average pairwise L2 distance from the prediction to the ground truth over all time steps. FDE measures the L2 distance from the predicted final position to the ground truth final position. It measures a model's ability for predicting the destination and is more challenging as errors accumulate in time.
Furthermore, we evaluate the most-likely prediction and the best prediction $@top10$, respectively.
Best prediction $@top10$ means among the 10 predicted trajectories with highest confidence, the one which has the smallest ADE and FDE is selected as the best. 

\begin{table}[t!]
\centering
\caption{\small{Experimental Results of different methods and models for mixed traffic. Unit is in meters and best values are highlighted in boldface. The smaller number is better. ``MCE" indicates MCENET and ``baseline" is the MCENET model without grouping or scene context. ``gp" stands for grouping context, ``hm" is for heat map, ``ap" is for aerial photograph and ``sm" denotes segmented map.}}
\label{tab:quantitativeresults}
\setlength{\tabcolsep}{7.3pt}
\renewcommand{\arraystretch}{0.9}
\begin{tabular}{|l|l|l|l|l|}
\hline
Data    & HBS                          & HC                                         & Gates3                      & Avg.  \\ 
\hline
\multicolumn{5}{|c|}{Most-likely predictions}   \\
\hline
\emph{S-LSTM}  & 1.67/3.03                    & 1.11/1.98                                  & 3.63/6.56                   & 2.14/3.86                          \\ 
\emph{S-GAN}   & 1.45/2.86                    & 0.97/1.67                                  & 2.98/5.42                   & 1.80/3.32                          \\ 
\emph{SS-LSTM} & 0.82/1.75                    & 0.49/0.79                                  & 1.61/\textbf{2.89}          & 0.97/1.81                          \\ 
\hline
\emph{Baseline} & 0.77/1.80                    & 0.49/0.80                                  & 1.54/3.04                   & 0.93/1.88                          \\ 
\emph{MCE+gp} & 0.77/1.80                    & \textbf{0.47}/0.77                         & 1.50/2.98                   & 0.91/1.85                          \\ 
\emph{MCE+hm} & 0.76/1.77                    & \textbf{0.47}/0.77                         & 1.52/2.99                   & 0.92/1.84                          \\  
\emph{MCE+hm+gp} & \textbf{0.71}/\textbf{1.68}  & 0.48/0.79                                  & 1.50/3.01                   & \textbf{0.89}/1.83                 \\ 
\emph{MCE+ap+gp} & 0.74/1.76                    & \textbf{0.47}/0.76                         & \textbf{1.48}/2.91          & 0.90/\textbf{1.81}                 \\ 
\emph{MCE+sm+gp} & 0.80/1.86                    & \textbf{0.47}/\textbf{0.75}                & \textbf{1.48}/2.98          & 0.92/1.86                          \\ 
\hline
\multicolumn{5}{|c|}{Best predictions ($@top10$)}   \\
\hline
\emph{Baseline}           & 0.57/1.28                    & 0.47/0.77                        & 1.19/2.26                   & 0.74/1.44            \\ 
\emph{MCE+gp}           & 0.56/1.26                    & 0.45/0.72                        & 1.17/2.34                   & 0.73/1.44            \\ 
\emph{MCE+hm}           & 0.57/1.26                    & 0.45/0.73                        & 1.20/2.24                   & \textbf{0.72}/1.41   \\ 
\emph{MCE+hm+gp}           & \textbf{0.55}/\textbf{1.20}  & 0.46/0.73                        & 1.18/2.26                   & 0.73/\textbf{1.40}   \\ 
\emph{MCE+ap+gp}           & \textbf{0.55}/1.25           & 0.45/0.72                        & 1.21/2.30                   & 0.74/1.42            \\ 
\emph{MCE+sm+gp}           & 0.60/1.32                    & \textbf{0.44}/\textbf{0.70}      & \textbf{1.15}/\textbf{2.22} & 0.73/1.41            \\ 
\hline
\end{tabular}
\end{table}
\subsection{Experiment Setting}
\label{sec:settings}
For interaction context, the occupancy has $8*8$ cells. In the DBSCAN cluster for group detection, $\epsilon$ is set to \SI{1.5}{m} and the co-existing rate is set to 0.9 empirically. For the neural networks, CONV-1D kernel size is set to 8; the CNN network has three layers with kernel sizes $(8, 4, 4)$; the hidden units of LSTMs are set to 128; and the dimension of the latent variables $z$ is 16. An Adam optimizer with a learning rate of 0.001 \cite{kingma2014adam} is applied for optimization. 

\begin{table}[ht!]
\begin{center}
\caption{\small{Experimental Results of different methods and models for pedestrians in eight time step and twelve time step prediction, respectively. The evaluation values for Social-GAN are the minimum values across all the sub-models reported from \cite{gupta2018social}. Unit is in meters and best values are in bold face.}}
\label{tab:quanresults}
\setlength{\tabcolsep}{10pt}
\renewcommand{\arraystretch}{1}
\begin{tabular}{|c|l|l|l|l|}
\hline
Model      & \multicolumn{1}{c|}{\emph{Social}} & \multicolumn{1}{c|}{\emph{Social}-} & \multicolumn{1}{c|}{\emph{SS-LSMT}} &                                   \multicolumn{1}{c|}{\emph{MCE}}\\ 
            & \multicolumn{1}{c|}{\emph{LSTM}}  & \multicolumn{1}{c|}{\emph{GAN}}     & \multicolumn{1}{c|}{\emph{hm}}   & \multicolumn{1}{c|}{\emph{hm+gp}}  \\ \hline
Data  & \multicolumn{4}{c|}{ADE (obs-8--pred-8 / obs-8--pred-12)} \\ \hline
Eth   & 0.70/1.09   & 0.60/0.81     & 0.66/0.80     & \textbf{0.58}/\textbf{0.75}     \\ \hline
Hotel & 0.55/0.86   & 0.48/0.67     & 0.32/0.47     & \textbf{0.23}/\textbf{0.37}     \\ \hline
Univ  & 0.36/0.61   & 0.36/\textbf{0.58}     & 0.54/0.78     & \textbf{0.35}/\textbf{0.58}     \\ \hline
Zara1 & 0.25/0.41     & 0.21/0.34     & 0.32/0.47     & \textbf{0.20}/\textbf{0.33}     \\ \hline
Zara2 & 0.31/0.52   & 0.27/\textbf{0.42}     & 0.39/0.62     & \textbf{0.23}/0.44     \\ \hline
Avg.  & 0.43/0.70   & 0.38/0.56     & 0.45/0.63     & \textbf{0.32}/\textbf{0.49}     \\ \hline
      & \multicolumn{4}{c|}{FDE (obs-8--pred-8 / obs-8--pred-12)}\\ \hline
Eth   & 1.45/2.41     & 1.19/\textbf{1.52}     & 1.23/1.57     & \textbf{1.10}/1.61     \\ \hline
Hotel & 1.17/1.91     & 0.95/1.37     & 0.55/0.90     & \textbf{0.38}/\textbf{0.68}     \\ \hline
Univ  & 0.77/1.31     & 0.73/1.22     & 0.99/1.50     & \textbf{0.70}/\textbf{1.18}     \\ \hline
Zara1 & 0.53/0.88     & 0.42/0.68     & 0.61/0.92     & \textbf{0.40}/\textbf{0.65}     \\ \hline
Zara2 & 0.65/1.11     & 0.54/0.84     & 0.67/1.19     & \textbf{0.44}/\textbf{0.79}     \\ \hline
Avg.  & 0.91/1.52     & 0.77/1.23     & 0.81/1.22     & \textbf{0.60}/\textbf{0.98}     \\ \hline
\end{tabular}
\end{center}
\end{table}

\subsection{Compared Methods and Ablative Models}
\label{subsec:compared_methods}
The proposed method is compared with the representative method S-LSTM \cite{alahi2016social} using deep learning technologies, and the most recent works S-GAN \cite{gupta2018social} and SS-LSTM \cite{xue2018ss}.
\begin{itemize}
    \item S-LSTM proposes a social pooling layer in which a rectangle occupancy gird is used to pool the existence of the neighborhood at each time step.
    After that, many following works \cite{lee2017desire,xue2018ss} adopt their social pooling layer for this task.
    \item S-GAN applies generative adversarial network for multiple future trajectories generation which is essentially different from previous works. It takes the interactions of all agents into account.
    \item SS-LSTM has an encoder-decoder structure. The encoder uses different LSTMs to encode the motion input, interaction input, and scene context.
\end{itemize}

To fully analyze the impact of each context, we carry out a series of ablation studies for the MCENET method. We treat the model as a baseline that removes grouping and scene context modules from MCENET. Then we add grouping and scene context from heat maps, aerial photographs and segmented maps to the MCENET models in parallel, see Tab.~\ref{tab:quantitativeresults}.

For the purpose of fare comparison, in mixed traffic trajectory prediction, we have implemented the codes of the above compared methods tested on Gates3~\cite{robicquet2016learning}, HBS~\cite{rinke2017multi} and HC~\cite{cheng2019pedestrian}; in pedestrian trajectory prediction, we do not use any extra image scene (\ie~aerial photographs or segmented maps) but only the trajectory data from ETH \cite{pellegrini2009you} and UCY \cite{lerner2007crowds}, so as to guarantee all the models have the same input data. The scene context is generated from the visible trajectories for both SS-LSTM and MCENET.

\begin{figure}[t!]
\centering
\begin{subfigure}{0.5\columnwidth}
\includegraphics[clip=true,trim=3pt 0pt 3pt 0pt, width=\columnwidth]{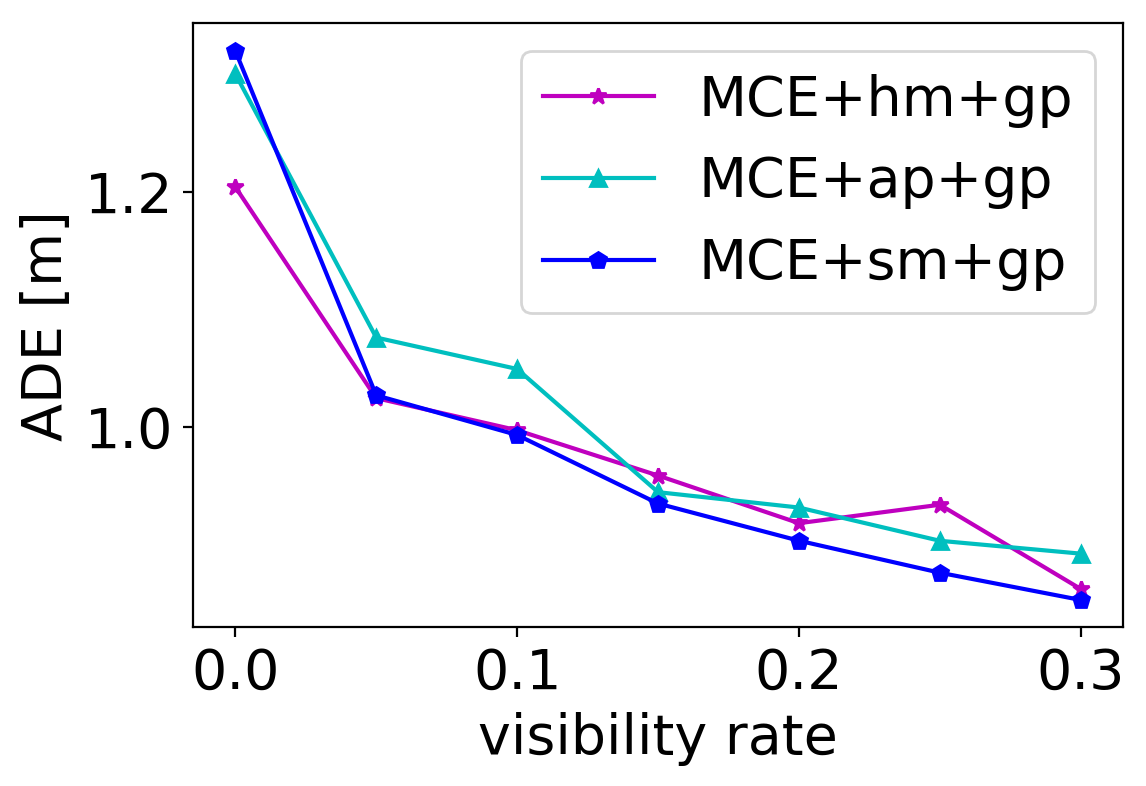}
\label{subfig:retrain_ade}
\end{subfigure}\hfill%
\begin{subfigure}{0.5\columnwidth}
\centering
\includegraphics[clip=true,trim=3pt 0pt 3pt 0pt, width=\columnwidth]{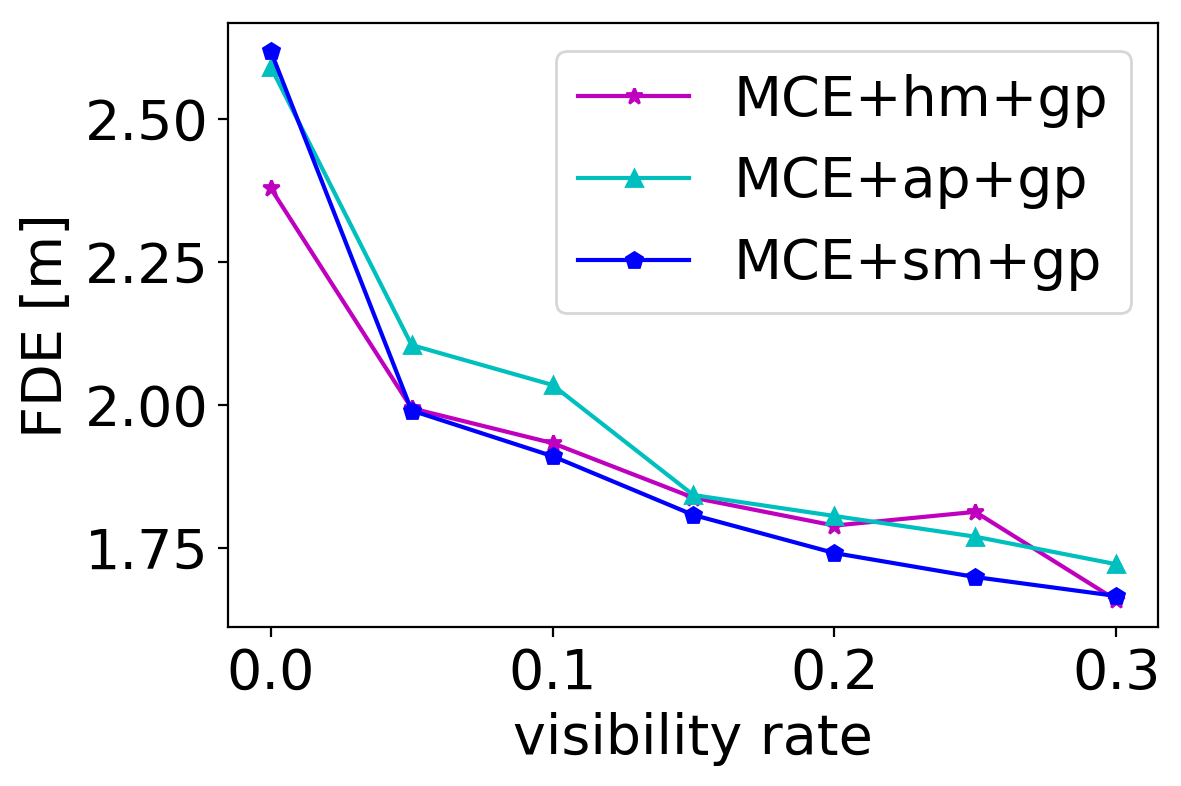}
\label{subfig:retrain_fde}
\end{subfigure}\hfill
\vspace{-1.8em}
\caption{\small{The performance for leave-one-out cross-validation and retraining with data taken from the target scene.}}
\label{fig:retrain}
\vspace{-12pt}
\end{figure}

\section{Results}
\label{sec:results}

\subsection{Quantitative Results}
\label{subsec:quan_results}
\textbf{Performance of Mixed Traffic Trajectory Prediction.} Tab.~\ref{tab:quantitativeresults} lists all the results measured by ADE/FDE of our method and the three compared stat-of-the-art models across all the mixed traffic datasets.
We can see that in most cases the MCENET models (with different scene contexts plus grouping context) outperform the other methods in predicting the most-likely trajectory. Only on Gates3, SS-LSTM slightly performs better regarding FDE. The better overall results reported by MCENET prove that grouping context and scene context are obviously helpful for trajectory prediction and MCENET is able to learn useful information from them effectively.
Meanwhile, the MCENET model reports much better results from the best predictions ($@top10$). 
It demonstrates that predicting multiple plausible trajectories is necessary and helpful to analyze how agent behave in the future.
It is worth noting that our baseline model reports comparable results with the state-of-the-art methods. It demonstrates that, our model is able to predict accurate future trajectory even only based on the history motion information.

To justify the impact of adopted contexts, several ablative models are also validated on our proposed MCENET method. The ablation study results are given in Tab.~\ref{tab:quantitativeresults}.
We can see that, the models utilizing scene and grouping contexts simultaneously have better overall performance than the models that partially consider grouping (MCE+gp) or scene context (MCE+hm). Regarding scene context, the model using heat maps and segmented maps perform comparable or better than the models using aerial photographs. 
It indicates that the motion prior of road agents in the heat maps is useful for predicting the future movements of mix-traffic road agents and the segmented maps provide explicit information about where is accessible respective to road agents' transport mode. 

After analyzing the positive impact of grouping and scene contexts, as well as their individual impact, we apply leave-one-out cross-validation to investigate the generalization ability of our model: predicting trajectories of heterogeneous agents in unseen space. We repeat this operation for each dataset and calculate the average performance. 
Fig.~\ref{fig:retrain} shows the average performance for the MCENET models that use different scene context. It can be seen that, with zero visibility rate of the target space ($0\%$ of the data from the test space is used for fine-tuning), the performance drops seriously compared with that what has been reported in Tab.~\ref{tab:quantitativeresults}.
It is a reasonable phenomenon, because different spaces have different scenarios. Scene information is an important factor for MCENET. The models trained in the other spaces have no knowledge about the scene information of the test space without fine-tuning. Therefore, the learned scene information does not match the one on the test set. We can see that with the increasing visibility rate, the performance of the models improve significantly. It also demonstrates that our model can be easily transferred to a new space through fine-tuning.

\textbf{Performance of Pedestrian Trajectory Prediction.} Table~\ref{tab:quanresults} shows the quantitative results measured by ADE and FDE for predicting eight time steps and twelve time steps side by side for pedestrians. 

Overall, MCENET outperforms the other approaches across all datasets measured by ADE. It marginally falls behind Social-GAN only on Zara2 for predicting eight time steps and on Eth for predicting twelve time steps regarding FDE. Meanwhile, the improvement margin on Hotel is even doubled for the MCENET model compared with Social-LSTM and Social-GAN.  

One interesting observation is that in the longer term prediction, SS-LSTM+hm outperforms Social-LSTM on Eth and Hotel. It indicates that the scene context is very important for trajectory prediction in long distance, as the environment may change when distance increases. On the other hand, when the image information is not available, heat maps manipulate the prior information of how different agents move in the past. They can be used as an alternative for scene context information.  

\begin{figure*} [t!]
\centering 
\begin{subfigure}{0.31\textwidth}
\includegraphics[clip=true,trim=0pt 0pt 0pt 0pt, width=\linewidth]{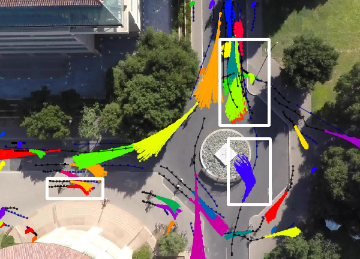}
\caption{\small{Baseline}}
\label{subfig:baseline}
\end{subfigure}\hfill
\begin{subfigure}{0.31\textwidth}
\includegraphics[clip=true,trim=0pt 0pt 0pt 0pt, width=\linewidth]{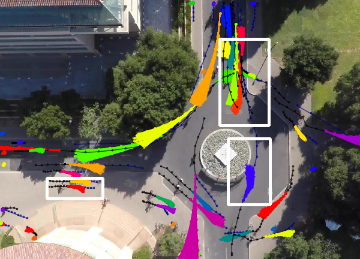}
\caption{\small{MCE+gp}}
\label{subfig:c+gp}
\end{subfigure}\hfill
\begin{subfigure}{0.31\textwidth}
\includegraphics[clip=true,trim=0pt 0pt 0pt 0pt, width=\linewidth]{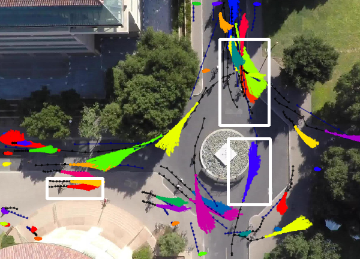}
\caption{\small{MCE+hm}}
\label{subfig:c+hm}
\end{subfigure}
\begin{subfigure}{0.31\textwidth}
\includegraphics[clip=true,trim=0pt 0pt 0pt 0pt, width=\linewidth]{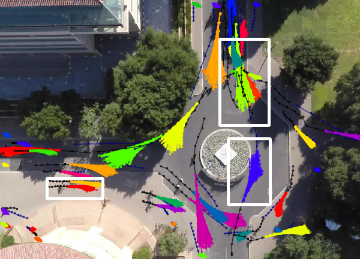}
\caption{\small{MCE+hm+gp}}
\label{subfig:c+hm+gp}
\end{subfigure}\hfill
\begin{subfigure}{0.31\textwidth}
\includegraphics[clip=true,trim=0pt 0pt 0pt 0pt, width=\linewidth]{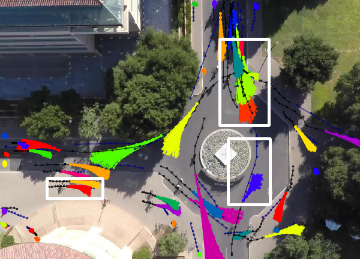}
\caption{\small{MCE+ap+gp}}
\label{subfig:c+ap+gp}
\end{subfigure}\hfill
\begin{subfigure}{0.31\textwidth}
\includegraphics[clip=true,trim=0pt 0pt 0pt 0pt, width=\linewidth]{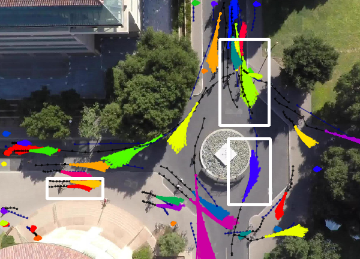}
\caption{\small{MCE+sm+gp}}
\label{subfig:c+sm+gp}
\end{subfigure}
\caption{\small{Examples of quantitative results from our method on the challenging Gates3 dataset. 
 MCE denotes MCENET and ``baseline" is the MCENET model that uses neither grouping nor scene context. ``gp" stands for grouping context, ``hm" is for heat maps, ``ap" is aerial photographs and ``sm" denotes segmented maps. Past trajectories are denoted in black and ground truth trajectories in purple. Different agents are denoted in different colors. Important differences are highlighted in white boxes.}
\label{fig:qualitativeresults}}
\end{figure*}

\begin{figure*} [t!]
\centering 
\begin{subfigure}{0.31\textwidth}
\includegraphics[clip=true,trim=0pt 20pt 0pt 0pt, width=\linewidth]{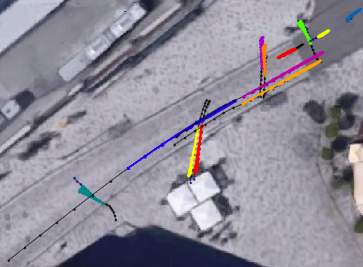}
\caption{\small{HBS}}
\label{subfig:hbs}
\end{subfigure}\hfill
\begin{subfigure}{0.31\textwidth}
\includegraphics[clip=true,trim=0pt 10pt 8pt 0pt, width=\linewidth]{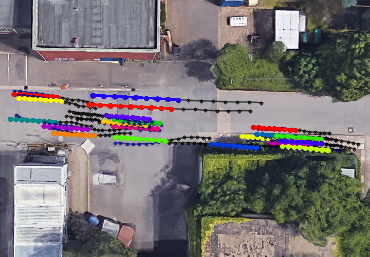}
\caption{\small{HC}}
\label{subfig:hc}
\end{subfigure}\hfill
\begin{subfigure}{0.31\textwidth}
\includegraphics[clip=true,trim=0pt 0pt 8pt 20pt, width=\linewidth]{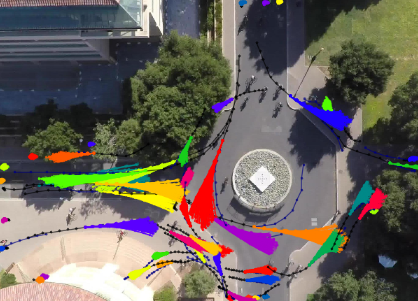}
\caption{\small{Gates3}}
\label{subfig:gates3}
\end{subfigure}
\caption{\small{Quantitative results of MCENET with grouping and segmented maps across different spaces in mixed traffic. Past trajectories are denoted in black and ground truth trajectories in purple.}}
\label{fig:qualitativeresults2}
\vspace{-12pt}
\end{figure*}

\subsection{Qualitative Results}
\label{subsec:qual_results}
Fig.~\ref{fig:qualitativeresults} and~\ref{fig:qualitativeresults2} show the qualitative results output by our MCENET models for multi-path trajectory predictions in mixed traffic. 
The impact of different contexts on predicting trajectories is visualized in Fig.~\ref{fig:qualitativeresults} for the very challenging space Gates3. Each sub-figure represents the utility of different contexts and the important differences are highlighted in white bounding boxes. 
It can be seen that using grouping context (Fig.~\ref{subfig:c+gp}) is helpful for converging the predicted multiple trajectories compared with the baseline (Fig.~\ref{subfig:baseline}).
In comparison between Fig.~\ref{subfig:c+hm} and Fig.~\ref{subfig:baseline}, using heat maps makes the prediction more accurate to the ground truth. For instance, the trajectories of the blue agent (the middle box) are incorrectly predicted toward the center of the cross by the baseline model. When the heat maps are used, the predicted trajectories are along the road and completely fit the ground truth. It indicates that the prior in the heat maps is important for predicting trajectory, especially in the scene with complex interactions. When the scene context of heat maps is integrated with grouping context (Fig.~\ref{subfig:c+hm+gp}) the prediction results are improved.

The comparison between the second row shows different impact of different kinds of scene context. We can see that the aerial photographs provide global visual context of the scene and improve the prediction compared with without using them (Fig.~\ref{subfig:c+gp}). It indicates that MCENET is able to extract useful information directly from the RGB image to help predict trajectories.
However, compared with the scene context of heat maps and segmented maps (Fig.~\ref{subfig:c+sm+gp}), its improvement is less, which is also in line with the quantitative results in Tab.~\ref{tab:quantitativeresults}. This is because the scene context in RGB images is implicit while heat maps and segmented maps provide explicit scene context. On the other hand, an RGB image is easier to be acquired than the heat maps and segmented maps, especially in complex scenes.
By comparing Fig.~\ref{subfig:c+sm+gp} with Fig.~\ref{subfig:c+hm+gp} we can see that, predicted trajectories with segmented maps have less divergence than the ones with heat maps.
This is because the segmented maps have strong constraints on how an agent should behave in the given scene while the heat maps have statistical prior on how an agent behaves. 

Fig.~\ref{fig:qualitativeresults2} demonstrates a full MCENET model with grouping and segmented maps across different spaces.
We can see that our method is able to predict the future trajectories of different agents (denoted by different colors) precisely by observing their history trajectories (in black). The predicted bunches of trajectory of any agent do not diverge much and are very close to the ground truth (covered by the prediction) in the less complex environment, \ie~HBS and HC. On the other hand, interactions between road agents are more complicated and each agent has more possibilities to choose their future paths in Gates3. Even though our method is able to predict the trajectory correctly, the predicted trajectories diverge more widely with further time step. It demonstrates the effectiveness of our model. It also proves that the ability of predicting multiple plausible trajectories is important in this task, because of the uncertainty of the future movements increasing in the longer term prediction.

\section{Conclusion}
\label{sec:con_fut}
We propose a novel framework MCENET for multi-path trajectory prediction of heterogeneous road agents in mixed traffic. The method incorporates scene context, interaction context and motion information to capture the variations of the future trajectories by learning a set of stochastic latent variables. Multi-path trajectories are predicted depending on the past information of target agent by introducing the stochastic latent variables. Particularly, the impact of three kinds of scene context are studied for this task. We demonstrate the efficacy of our method on several complicated real-world scenarios and clear improvement over other recent state-of-the-art approaches.

\section{Acknowledgments}
This work is supported by the German Research Foundation (DFG) through the Research Training Group SocialCars (GRK 1931) and grants COVMAP (RO 2497/12-2).

\bibliographystyle{ieee_fullname}
\bibliography{mcenet}

\end{document}